PAPER • OPEN ACCESS

# Prediction of Solar Radiation Using Artificial Neural Network

To cite this article: Shahriar Rahman *et al* 2021 *J. Phys.: Conf. Ser.* **1767** 012041

View the article online for updates and enhancements.

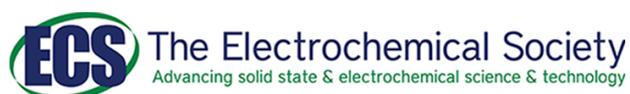
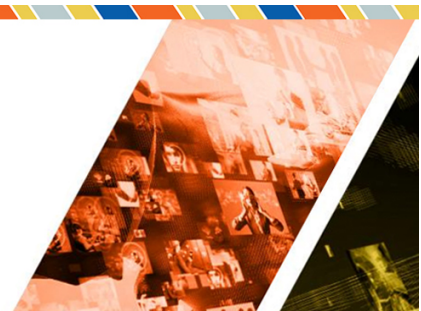

# Prediction of Solar Radiation Using Artificial Neural Network

**Shahriar Rahman[1], Shazzadur Rahman[2] and A K M Bahalul Haque[3]**

[1,2,3] North South University, Bangladesh

[1] shahriar.rahman10@northsouth.edu
[2] shazzadur.rahman01@northsouth.edu
[3] bahalul.haque@northsouth.edu

**Abstract** – Most solar applications and systems can be reliably used to generate electricity and power in many homes and offices. Recently, there is an increase in many solar required systems that can be found not only in electricity generation but other applications such as solar distillation, water heating, heating of buildings, meteorology and producing solar conversion energy. Prediction of solar radiation is very significant in order to accomplish the previously mentioned objectives. In this paper, the main target is to present an algorithm that can be used to predict an hourly activity of solar radiation. Using a dataset that consists of temperature of air, time, humidity, wind speed, atmospheric pressure, direction of wind and solar radiation data, an Artificial Neural Network (ANN) model is constructed to effectively forecast solar radiation using the available weather forecast data. Two models are created to efficiently create a system capable of interpreting patterns through supervised learning data and predict the correct amount of radiation present in the atmosphere. The results of the two statistical indicators: Mean Absolute Error (MAE) and Mean Squared Error (MSE) are performed and compared with observed and predicted data. These two models were able to generate efficient predictions with sufficient performance accuracy.

## 1. Introduction

With the rise in technological advancements in our digital modern world, comes the rise in demand for electricity [1]. Solar Panels have become one of the most used devices for electricity production as it has become more affordable and more efficient than ever [2]. Solar radiation data is significant in various sectors such as in conversion and generation of energy from sunlight, water heating, water distillation and meteorology [3, 4]. Many solar technologies started to take full advantage of using solar radiation as a foundation for producing electrical energy. Furthermore, insolation energy has proven to be very valuable in other sectors as well, such as agricultural sectors and rainfall measurement and detection. On the other hand, it also can have minor negative effects such as: Radiation Exposure, UV and infra-red rays and climate change, thus making it more essential to analyze the radiation data.

　　To address these problems and achieving proper radiation data for energy conversion and other useful applications, many studies in the literature explored using various techniques of machine learning to precisely obtain the necessary data. Generally, solar radiation data can be observed as a time series produced by a random and stochastic process [5]. As a result, precise mathematical modeling is necessary for efficient generalization. Hence, by using the historical data sample, it can be mathematically interpreted as a conditional expectation by using a precise model.

　　In this paper, we propose two models, one constructed using Mean Squared Error (MSE) and another using Mean Absolute Error (MAE). These two models are used inside deep neural network architecture, and thus, contributing a considerable capacity to learn the representations mentioned above. Having a pretty decent amount of labeled data with different types of features for this task, it makes it fairly easy to train a network with a large number of parameters. However, there is a slight problem, the data distribution is quite imbalanced. Even with the proper structure and network architectures, the deviations of data make it very difficult to obtain improved generalization. For this reason, we decided to use the more optimized path to handle this situation, which means more training time and computational cost in exchange for better prediction results. These disadvantages are then tackled and minimized with efficient tuning and using a graphics processing unit (GPU) for training procedures.

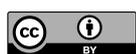

## 2. Literature Review

In this section, different types of related state-of-the-art models will be discussed. As mentioned above, the prediction of solar radiation models has random, statistical and deep learning methods. One of the few statistical models such as the exponentially weighted moving average, also known as EWMA in short, is most commonly used. Other models such as Artificial Neural Network using regression techniques are used from a given set of past samples with the expected conditions, while Markov and Auto-Regressive models are used for unsupervised training and predictions.

### 2.1. Exponentially Weighted Moving Average

The EWMA is one of the most popular algorithms used for solar prediction models. What EWMA does is that it divides a single day into a various number of constant or fixed time slots [6]. The interval of the time slots can range between 30 minutes to an hour and the number of time slots is considered N. The normal assumption is that the energy accumulated on a specific day is considered to be the same on the previous days, given that the time in which the data is collected on both days is similar. Thus, the following equation can be written such as:

$$E(d, N) = \alpha * E(d - 1, N) + (1 - \alpha) * H * (d - 1, N) \quad (1)$$

Here, E (d, N) represents the estimated energy for the current day, where d signifies the current date and N stands for the number of time slots. Similarly, E (d - 1, N) represents the estimated energy for the previous day and H means the last energy harvested. Lastly alpha stands for weighting factor and it ranges from 0 to 1. Alpha plays an important factor that is used to move the weighted average, which otherwise would be inefficient and inconsistent.

### 2.2. Weather Conditioned Moving Average

Also known as WCMA in short, the Weather Conditioned Moving Average is another statistical algorithm like EWMA [7, 8]. Unlike EWMA, however, the algorithm does not consist of any weighted average. So, instead of taking the weighted average into account, it includes the energy accumulated in the last time slot. Therefore, if energy is to be estimated for the current slot, then the energy estimation of the previous slot needs to be taken into account, which gives the following formula:

$$E(d, N + 1) = \alpha * E(d, N) + (1 - \alpha) * M * (d, N + 1) * GAP(d, n, k) \quad (2)$$

Here, M gives the average of (N+1)th slots in the previous d number of days. The harvested energy collected from the last slot is defined by E (d, N) and similarly alpha is the weighted factor. GAP (d, n, k) represents the comparison of the current solar radiation condition with the previous days.

$$GAP(d, n, k) = \frac{V.P}{\sum P} \quad (3)$$

### 2.3. Artificial Neural Networks

Another approach includes using a Machine Learning model for generalization. By using Artificial Neural Network (ANN), the models are trained to generalize from the available historical dataset from which it is capable of mining non-linear modeling relationships. Assuming the structures and the core features of the ANN are satisfactory, this type of model can understand the pattern and extract all relevant relationships as a result. Mellit, Benghanem and Bendekhis [3] proposed a model which will take in parameters such as regular weather forecasting data like daily temperature and sunshine duration to properly get predicted radiation values. Sfetsos and Coonick [9] proposed an adaptive neuro fuzzy inference and ANN models in which it uses temperature, pressure, wind speed and wind direction as an additional parameter.

*2.4. Unsupervised Models*

Sometimes, the conditional expectations from past samples are not known, thus find the optimal solution is hard to control. For this reason, and also the fact that these are generally non-linear, the methods for designing a model to predict from such samples are tackled using Auto-Regressive Models [10], Markov Models [11] and Auto-Regressive moving average models, or ARMA. These are used to generate statistical assumptions that provide a good solution to such stated problems.

**3. Data Set**

This section consists of information about the data set collected and used in order to find proper solutions to the stated problem.

*3.1. Data Collection*

The meteorological data consists of air pressure, time, humidity, wind speed, daily temperature, wind direction and global solar radiation. These data were recorded by a meteorological station from the Hawai'i Space Exploration Analog and Simulation weather station. Also known as HI-SEAS. The time period of the data collected is for four months (September through December, 2016) between Mission IV and Mission V. HI-SEAS is an environment located on a remote site on the Mauna Loa side of the saddle area on the Big Island of Hawaii in around 8100 feet above sea level [12].

*3.2. Data Analysis*

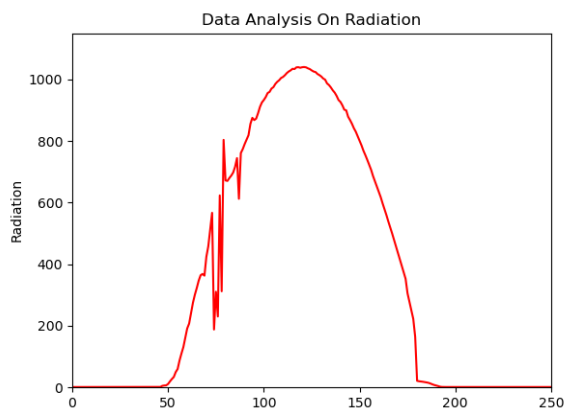

**Figure 1.** Graph of a global daily solar radiation sequence in watt per square meter, W/m$^2$.

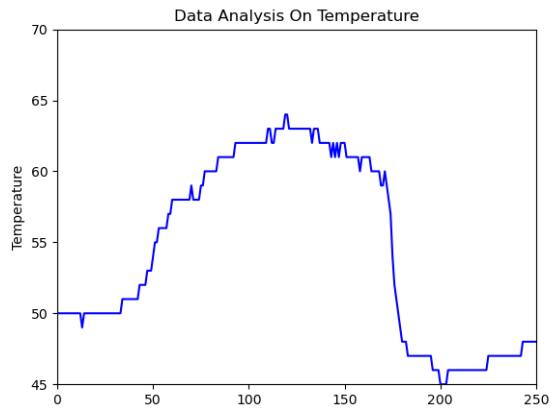

**Figure 2.** Graph of an average day temperature in Celsius sequence.

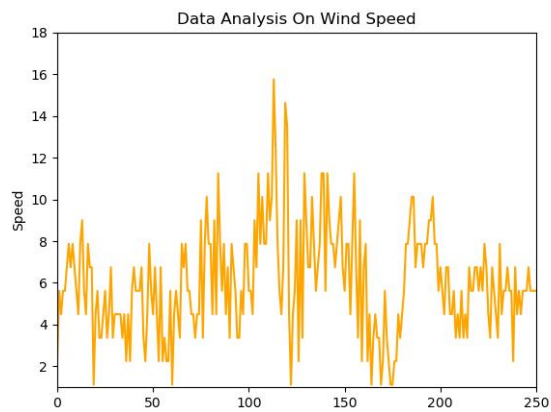

**Figure 3.** Single-day wind speed in m/s Graph (might differ depending on seasons).

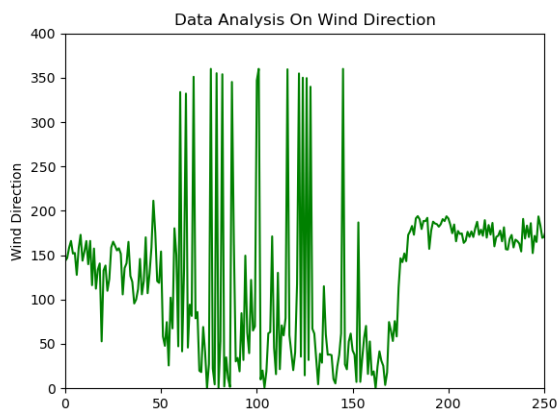

**Figure 4.** Wind direction sequence in a single day.

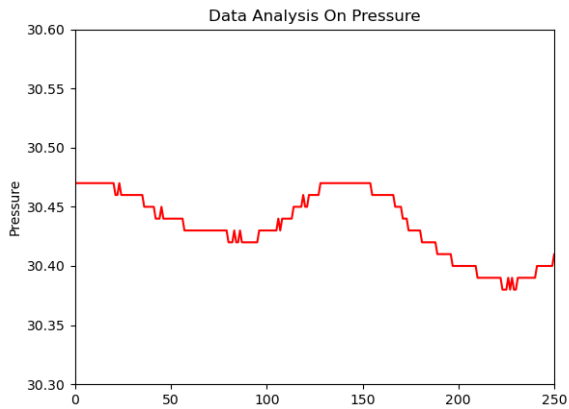

**Figure 5.** Graph of a typical example of a global daily atmospheric pressure in Pascals (mostly constant).

**Table 1.** Table consisting of monthly average hourly radiation, temperature and pressure.

| Months | Radiation (W/m$^2$) | Temperature (Celsius) | Pressure (Pascals) |
| --- | --- | --- | --- |
| September | 222.36 | 54.10 | 30.41 |
| October | 237.34 | 52.79 | 30.43 |
| November | 215.61 | 50.21 | 30.41 |
| December | 138.43 | 47.95 | 30.37 |

*3.3. Data Observations*

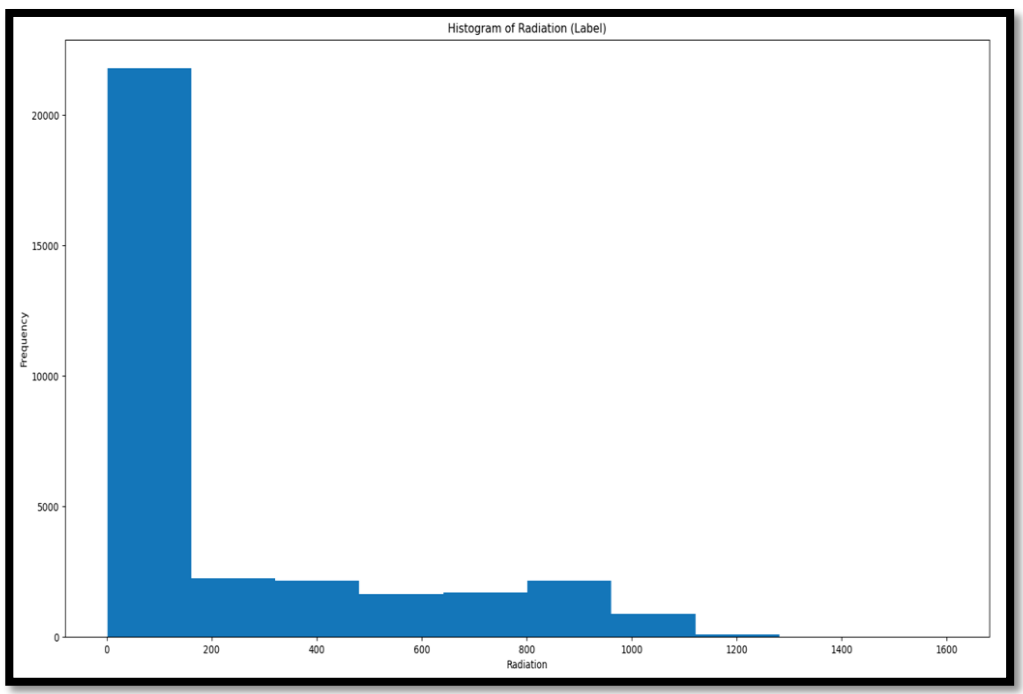

**Figure 6.** Histogram of the radiation values.

Upon observing the histogram and all the previous graphical representations of the data collected, it can be concluded the data set has a severe imbalance. From the Radiation Histogram, we can perceive that the difference in Solar Radiation values is immensely high, ranging from 0.15 to over 1000.00 W/m$^2$.

Similarly, the temperature, wind humidity and pressure also have a high level of variance, which would ultimately affect the predictive ability of the model. Due to high inconsistency, the sensible approach would be to focus on a cost-sensitive learning or hyperparameter values rather than safe computational load. The only disadvantage would be that it would require more time and computational load to train the training and validation sets.

## 4. Developed Model

In this section, the paper describes the process and architecture of the implementation constructed in order to make accurate predictions.

### 4.1. Model Flow Chart

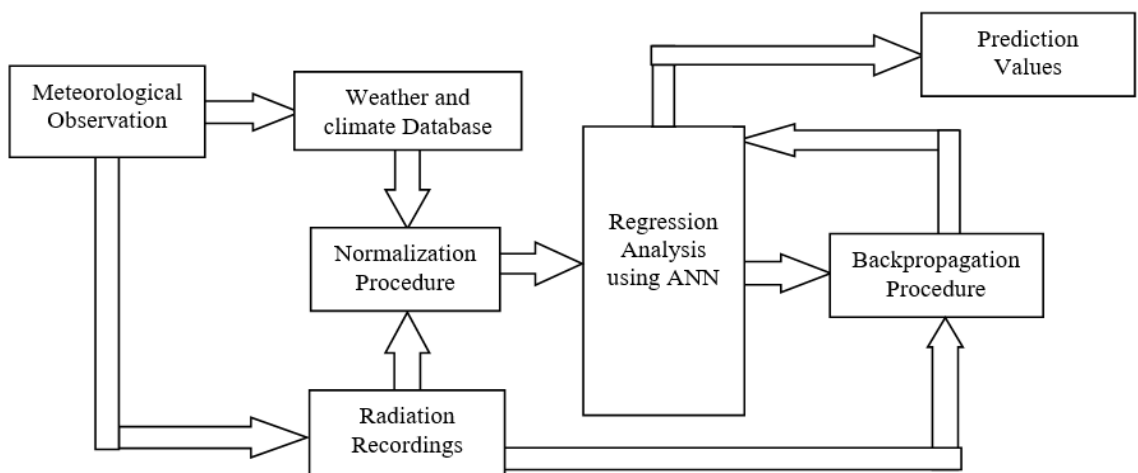

**Figure 7.** Procedural Diagram of the Predictive ANN Model.

### 4.2. Implementation

ANN Model: A Multi-Layered Perceptron (MLP) is implemented by using a feed-forward back-propagation algorithm.

ReLu Activation Function: One of the most widely used activation functions in deep learning models, the Rectified Linear Unit or ReLu is a function that returns zero from the input if it has any values that is not positive, however, it returns the exact value for any positive inputs [13]. Thus, it can be written as: $f(x) = max(0, x)$ 　　　　　　　　　　　　　　　　　　　　　　　　　　(4)

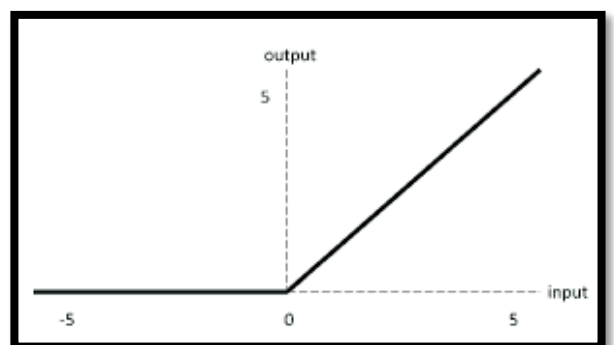

**Figure 8.** A Graph of a ReLu Activation Function.

Computation: The artificial neural network computation can be separated into two stages: Learning stage and Testing stage. The network in the learning stage makes errors and learns from it by adjusting the weights and biases to properly predict the correct or actual class label of the input tuples. Mathematically, the function of the processing elements can be expressed as:

$$z(x,w) = \sum_{i=1}^{n}(w_i \cdot x_i + b) \quad (5)$$

Where w is synaptic weights and b is the bias value. They are learnable parameters, which act as a catalyst to the model's learning capabilities.

MSE Loss Function: For most regression problems, the Mean Squared Error, also known as the MSE loss is the default loss function and for this problem, this is no different [14]. For a single example, MSE is calculated by determining the difference in error between the calculated output by the model and the output prediction that is provided. We, then, square the obtained error.

$$MSE = \frac{1}{N} + \sum_{i=1}^{n}(y_i - \hat{y}_i)^2 \quad (6)$$

Here, N is the number of samples, y is ground truth and $\hat{y}$ is the calculated output.

Optimizer: An Adam optimization algorithm is used in this model which is an addition to the SGD (stochastic gradient descent) and currently is one of the best optimizers [15]. Adam inherits from RMSProp and Gradient Descent with Momentum [16]. Equation (7) displays the backpropagation parameter adjustment using Adam optimizer.

$$w_t = w_{t-1} - \alpha \frac{m_t}{\sqrt{v_t} + \epsilon} \quad (7)$$

Where w is the weights of the learning model, α is the learning rate $m_t$ and $v_t$ are moving averages.

*4.3. Architecture*
Since, the ANN model requires high computational cost, we will be training it using a GeForce RTX 2080 Ti GPU developed by NVIDIA, which should serve quite well. Before initializing the model training procedure using GPU, we would normalize the features using Batch normalization.

$$norm = \frac{x_i - \mu}{\sigma} \quad (8)$$

In this equation, $x_i$ is the sample feature, μ is the mean value and $\sigma$ represents standard deviation. Generally, an additional notation Є is used to ensure that the denominator does not reach zero. In our case, we ignored the use of Є, simply because we would not need it as our denominator did not reach zero.

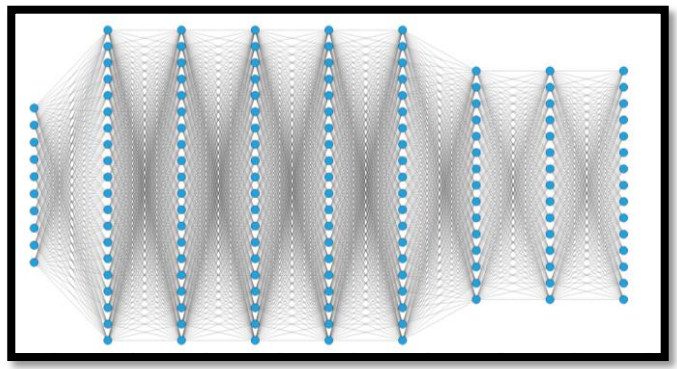

**Figure 9.** The Proposed ANN Model Architecture.

The architecture consists of 9 layers. The first layer, with a depth of 10 units, the next 5 layers with a depth of 20 units, then decreasing to 15 units for the last 3 layers. In the last 2 layers of the model, we applied Ridge Regression, also known as L2 Regularization [17]. The ridge regression adds a coefficient penalty term known as the "squared magnitude" to the loss function. As a result, it helps to solve the overfitting problem.

$$L(x,y) \equiv \sum_{i=1}^{n}(y_i - h_\theta(x_i))^2 + \lambda \sum_{i=1}^{n}|\theta_i| \qquad (9)$$

The values used for both kernel and bias regularizer is set to 0.009. Also, the total number of trainable parameters in the model are 2,801.

**Table 2.** A table consisting of all model parameters.

| Layers | Activation Shape | Number of Parameters |
|---|---|---|
| Input to Dense layers | (10, 1) | 90 |
| Dense 1 | (20, 1) | 220 |
| Dense 2 | (20, 1) | 420 |
| Dense 3 | (20, 1) | 420 |
| Dense 4 | (20, 1) | 420 |
| Dense 5 | (20, 1) | 420 |
| Dense 6 | (15, 1) | 315 |
| Dense 7 | (15, 1) | 240 |
| Dense 8 | (15, 1) | 240 |
| Dense 9 | (1, 1) | 16 |
| Total Parameters | | 2,801 |

*4.4. Data Partitions*
We split our training set into 3 partitions. First, we split the set into 70/30, one half for training and the other half for both validation and testing. Then, we split the 30% of the set into 2 parts, 50% for validation and another 50% for the testing set. Therefore, the combined split of our data set is 70/15/15 where 70% is for training, 15% for validation and another 15% for the testing set.

Furthermore, we used a batch of 512 as it showed more consistent results. We also used a technique called early stopping callback, to properly diagnose and identify the number of ideal epochs required for the model to have an appropriate fitting. After playing around with it, we discovered that any value more than 160 seems to overfit, hence we initialized the number of epochs to be 152.

**5. Results and Discussion**
Initially, the model consisted of 8 layers with a regularizer values of 0.01 for both bias and kernel. After running the simulation, it is observed that better performance is accomplished by adding an extra layer with 15 depth and using a regularizing value of 0.009. We also faced some inconsistencies while using the Adam optimizer, even though it performed far better than other optimizers.

*5.1. Optimizations*
However, we managed to locate the inconsistency in Adam and used a technique to find a workaround.

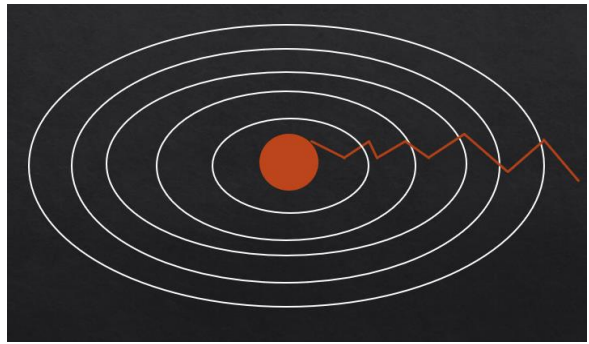

**Figure 10.** Convergence of Adam.

Adam is a combination of RMSProp and Momentum [18]. RMSProp helps reduce the vertical oscillation where Gradient descent with momentum adds momentum towards the horizontal direction, which is good because it prevents overshooting. However, even though we want to speed up the learning process at the start because it speeds up the learning process, it also needs to slow down after a while otherwise it would have difficulties converging. This is why we introduced another optimization technique known as learning rate decay.

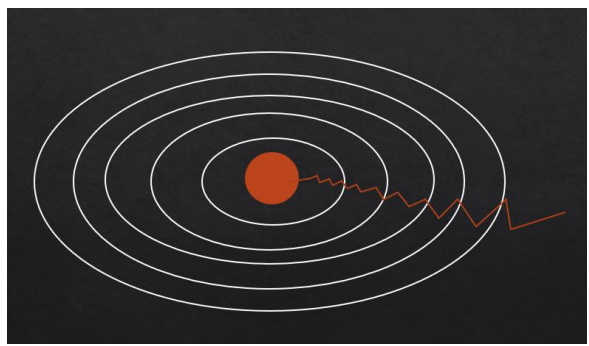

**Figure 11.** Convergence of Adam with decay.

Learning rate decay allows the model to train at a much faster speed at the start, however, after taking some steps, it starts to reduce the speed of the learning process. Thus, giving it more time to converge properly.

*5.2. Error Comparison*
As a result, after combining Adam optimizer with learning rate decay, we obtained a satisfactory predictive result, with a Training Loss of only around 0.0726 and a Validation Loss of around 0.0995.

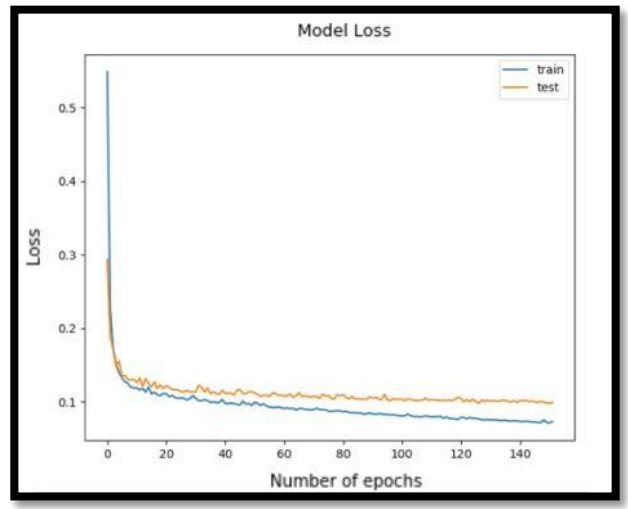

**Figure 12.** Training vs Validation Graph (MSE).

We also observed and compared the results of unseen or test predicted values with the actual test values. Most predicted values are pretty close to the actual value with a small average deviation value of approximately 20 from the actual value. Sometimes, the results have an exact match with the actual value. Thus, we are pleased with the results.

There is one caveat though, whenever we tried using the Mean Absolute Error Loss function instead of Mean Squared Error, the results vary by a small margin. Firstly, the loss seems to increase for the training procedure. But the model is able to generalize more consistently.

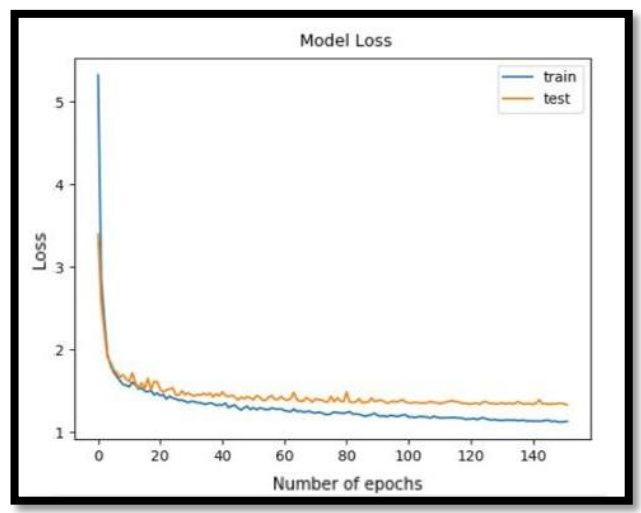

**Figure 13.** Training vs Validation Graph (MAE).

The Training Loss obtained by using MAE is 0.1118 and the Validation Loss of 0.1331 but the predicted value is more accurate than that of MSE. However, we could not find any logical reasoning behind this occurrence other than the fact that using MAE reduces overfitting by a small margin.

## 6. Conclusion
This paper explains the procedure of modeling and generalizing hourly solar radiation from an observed dataset, which is statistically close to actual values. The values obtained with the use of deep learning and an artificial neural network trained with historical data values and it shows that the developed tools allow the estimation of values of radiation accurately. With this information, many potential applications can benefit from it. Applications such as the power output of a photovoltaic production center can be forecasted. It can also be used to detect any environmental and weather-related applications.

The advantage of this model is that it can forecast and estimate a sequence of hourly solar radiation data correctly from weather forecasting data which can be collected fairly easily. However, one drawback of this model is that there is minor overfitting with the validation set, which, after spending a significant portion of time, could not be able to be reduced any further. Only once, the overfitting was able to be fixed but it was at the cost of a larger error, and also the generalization was not as good as our current two models.